\documentclass{article}
\usepackage[margin=1.0in]{geometry} 
\usepackage[utf8]{inputenc}
\usepackage[bitstream-charter]{mathdesign}

\usepackage{cite}
\usepackage{graphicx}
\usepackage{subfigure}
\usepackage[colorlinks=true, linkcolor=black, citecolor=blue, urlcolor=blue]{hyperref}
\usepackage{blindtext}
\usepackage{changepage}
\usepackage{breqn}
\usepackage[plain]{algorithm2e}
\usepackage[english]{babel}
\usepackage{amsthm}

\usepackage{graphicx}
\usepackage{float}
\usepackage{bbm}
\usepackage{marginnote}
\usepackage{qtree}
\usepackage{footmisc}
\usepackage[affil-it]{authblk}
\usepackage{epigraph}
\usepackage{makeidx}
\usepackage{bbding}
\usepackage{wasysym}

\usepackage{multirow}
\usepackage{makecell}

\usepackage{footnote}

\usepackage{fancyhdr}

\makeindex

\let\oldindex\index
\newcommand{\magic}[1]{%
    \oldindex{#1}%
}

\newcommand{\matr}[1]{\mathbf{#1}}  

\title{Interpretability and Explainability:\\ A Machine Learning Zoo Mini-tour}

\author{Ri\v{c}ards Marcinkevi\v{c}s\href{mailto:ricards.marcinkevics@inf.ethz.ch}{\textsuperscript{\Envelope}}%
    \thanks{\href{mailto:ricards.marcinkevics@inf.ethz.ch}{\textsuperscript{\Envelope} ricards.marcinkevics@inf.ethz.ch}}, Julia E. Vogt}
    \affil{ETH Zurich, Department of Computer Science, \\ Institute for Machine Learning}

\date{December, 2020}

\fancypagestyle{plain}{%
  \fancyhf{} 
  \fancyfoot[L]{\footnotesize This is a preprint version of the 2023 WIREs Data Mining and Knowledge Discovery article. \\ \textbf{Published version:} Marcinkevi\v{c}s, R., \& Vogt, J. E. (2023) Interpretable and explainable machine learning: A methods-centric overview with concrete examples. \textit{WIREs Data Mining and Knowledge Discovery}, e1493. doi: \href{https://doi.org/10.1002/widm.1493}{10.1002/widm.1493}}
  
}

\begin{document}

\renewcommand{\abstractname}{\vspace{-\baselineskip}}   
\maketitle

\begin{abstract}
\noindent In this review, we examine the problem of designing interpretable and explainable machine learning models. Interpretability and explainability lie at the core of many machine learning and statistical applications in medicine, economics, law, and natural sciences. Although interpretability and explainability have escaped a clear universal definition, many techniques motivated by these properties have been developed over the recent 30 years with the focus currently shifting towards deep learning methods. In this review, we emphasise the divide between interpretability and explainability and illustrate these two different research directions with concrete examples of the state-of-the-art. The review is intended for a general machine learning audience with interest in exploring the problems of interpretation and explanation beyond logistic regression or random forest variable importance. This work is not an exhaustive literature survey, but rather a primer focusing selectively on certain lines of research which the authors found interesting or informative.
\end{abstract}

\section{Interpretability, Explainability, and Intelligibility\label{sec:intro}}

Interpretable and explainable ML techniques emerge from a need to design \emph{intelligible} machine learning systems, \emph{i.e.~}ones that can be comprehended by a human mind, and to understand and explain predictions made by \emph{opaque} models, such as deep neural networks \cite{Krizhevsky2017} or gradient boosting machines \cite{Mason1999, Friedman2001}. The early research on interpretable machine learning dates back to the 1990s \cite{Rudin2019} and often does not refer to terms like ``\emph{interpretability}'' or ``\emph{explainability}'', not to mention that many classical statistical models can be deemed interpretable.

In general, there is no agreement within the ML community on the definition of \emph{interpretability}\magic{interpretability} and the \emph{task of interpretation} \cite{DoshiVelez2017, Lipton2018}. For example, Doshi-Velez and Kim define interpretability of ML systems as ``\emph{the ability to explain or to present in understandable terms to a human}'' \cite{DoshiVelez2017}. This definition clearly lacks mathematical rigour \cite{Lipton2018}. Nevertheless, the notion of interpretability often depends on the domain of application \cite{Rudin2019} and the target \emph{explainee} \cite{Carvalho2019}, \emph{i.e.} the recipient of interpretations and explanations, therefore, an all-purpose definition might be infeasible \cite{Rudin2019} or unnecessary. Other terms that are synonymous to interpretability and also appear in the ML literature are ``\emph{intelligibility}''\magic{intelligibility} \cite{Lou2012, Caruana2015} and ``\emph{understandability}'' \cite{Lipton2018}. These concepts are often used interchangeably.

Yet another term prevalent in the literature is ``\emph{explainability}'',\magic{explainability, XAI} giving rise to the direction of \emph{explainable artificial intelligence} (XAI) \cite{DARPAXAI}. This concept is closely tied with interpretability; and many authors do not differentiate between the two \cite{Carvalho2019}. Doshi-Velez and Kim \cite{DoshiVelez2017} provide a definition of \emph{explanation} that originates from psychology: ``\emph{explanations are... the currency in which we exchange beliefs}''.  In \cite{Rudin2019}, Rudin draws a clear line between interpretable and explainable ML: \emph{interpretable ML} focuses on designing models that are \emph{inherently interpretable}; whereas \emph{explainable ML} tries to provide \emph{post hoc} explanations for existing \emph{black box} models\magic{black box}, \emph{i.e.} models that are incomprehensible to humans or are proprietary \cite{Rudin2019}. For the sake of convenience, in this review, we adhere to the distinction proposed by Rudin \cite{Rudin2019}. In \cite{Lipton2018}, Lipton stresses the difference in questions the two families of techniques try to address: interpretability raises the question ``\emph{How does the model work?}''; whereas explanation methods try to answer ``\emph{What else can the model tell me?}''.

Since there is no general definition of either interpretability or explainability, researchers have elicited various \emph{desiderata},\magic{desiderata for interpretability} diverse and often contradicting, to provide motivation for different techniques. Below we summarise, based on \cite{DoshiVelez2017}, \cite{Lipton2018}, and \cite{Carvalho2019}, a few of the goals that could be achieved with interpretability:
\begin{itemize}
    \item \textbf{Trust} is often mentioned in association with interpretability \cite{DoshiVelez2017, Lipton2018}. Interpretability can be a prerequisite for a trustable ML system. Lipton \cite{Lipton2018} decomposes trust into knowing ``\emph{how often a model is right}'' and ``\emph{for which examples it is right}''.
    \item \textbf{Causality} might be desirable when we want to leverage an interpretable model or an explanation method to generate hypotheses about causal relationships between variables \cite{Lipton2018}.\magic{causality} Therefore, it is often desirable for the technique to pick up causal relationships \cite{Carvalho2019}, rather than mere associations. Such formulation of interpretability usually requires solving the problem of the \emph{observational causal discovery} \cite{Mooij2016}.\magic{observational causal discovery}
    \item \textbf{Reliability / Robustness / Transferability} \cite{DoshiVelez2017, Lipton2018, Carvalho2019} posit that ML systems should be resistant to noisy inputs and (reasonable) \emph{domain shifts}. An example wherein interpretability facilitates domain adaptation is given by Caruana \emph{et al.} in \cite{Caruana2015} who use additive models for pneumonia risk prediction, exhibit and alleviate unwanted confounding in the dataset. In general, this direction of research is strongly related to the problems of \emph{domain adaptation} and \emph{transfer learning} \cite{Kouw2018}.\magic{domain adaptation}
    \item \textbf{Fairness} is important when ML algorithms are incorporated into decision-making, \emph{e.g.} social, economical or medical.\magic{fairness} Interpretations and explanations can be instrumental in exposing demographic and other biases in algorithms and datasets \cite{Lipton2018, Carvalho2019}. Needless to mention, this is not a straightforward task, especially, when an explanation method is not \emph{faithful} w.r.t. the original predictive model \cite{Rudin2019}, \emph{i.e.} does not really reflect inner workings of the model.
    \item \textbf{Privacy} \cite{DoshiVelez2017, Carvalho2019} can be of concern in systems relying on sensitive personal data.\magic{privacy preservation} Similarly to fairness, interpretations and explanations can help to understand if user \emph{privacy is preserved}.
\end{itemize}

In the remainder of this review, we briefly discuss a need for interpretable and explainable machine learning techniques giving examples from several application areas (see Section~\ref{sec:relevance}). We provide an overview of evaluation methods of interpretability and explainability  (see Section~\ref{sec:eval}). The main body of this paper (see Section~\ref{sec:methodology}) outlines a partial taxonomy of techniques for interpretable (see Section~\ref{sec:interpretablemodels}) and explainable (see Section~\ref{sec:explainable}) ML with examples of several recent advancements. Summary of reviewed methods is provided in Tables~\ref{tab:interpretmethodsummary} and \ref{tab:explmethodsummary}. We conclude with a brief summary in Section~\ref{sec:summary}.

\section{Relevance\label{sec:relevance}}

It is natural to question usefulness of interpretable and explainable ML, especially, given a widespread belief that there exists a \emph{trade-off between accuracy and interpretability}\magic{accuracy-interpretability trade-off} \cite{Rudin2019}. \emph{Why would a designer of an ML system consider sacrificing performance for the sake of transparency?} First, it is important to note that there are many cases when interpretability is not necessary at all, in particular, when the studied problem is well-known and does not have strong consequences \cite{DoshiVelez2017}, \emph{e.g.} mail sorting, movie recommendation etc. Second, the perceived accuracy-interpretability trade-off may not necessarily exist in many datasets \cite{Rudin2019}. Doshi-Velez and Kim \cite{DoshiVelez2017} claim that a natural need for interpretability arises when a formalisation of the problem is \emph{incomplete},\magic{incompleteness} thus, making direct optimisation and validation impossible. They provide a few examples of such situations \cite{DoshiVelez2017}, we extend their list with references to some concrete instances:
\begin{itemize}
    \item \textbf{Scientific Understanding}: we often want to extract (scientific) knowledge from the data, and interpretations and explanations might be the best way to achieve this \cite{DoshiVelez2017}. Interpretability can be instrumental in \emph{exploratory data analysis} and discovery. This application of ML is only emerging. For example, interpretable support vector machines have been used for discovering new phases of Kitaev materials \cite{Liu2020}; in quantum chemistry, neural networks have been applied to predict the quantum mechanical wavefunction \cite{Schuett2019}, allowing for its analytical differentiable representation. In computational linguistics, Pimentel \emph{et al.} \cite{Pimentel2019} have leveraged NLP models alongside with an information-theoretic approach to quantify the relationship between form and meaning of words and test the long-standing Saussure's hypothesis. These are just few examples of machine-learning-assisted scientific discovery, a comprehensive survey by Raghu and Schmidt \cite{Raghu2020} contains a plethora of scientific deep learning applications.
    \item \textbf{Safety}: interpretability can assist developers in understanding when an ML system can or will fail \cite{DoshiVelez2017}. We have already mentioned the pneumonia prediction example from Caruana \emph{et al.} \cite{Caruana2015}, where interpretability allowed revealing confounding between a predictor variable and the response. Another noteworthy example is the \emph{Manifold} -- an in-house visualisation and debugging system for ML models developed at Uber \cite{Carvalho2019, Li2019}.
    \item \textbf{Ethics}: it might be difficult to embed all ethical and fairness constraints into an ML system \cite{DoshiVelez2017}. Likewise with safety, interpretations and explanations may help to `debug' unfair models. For example, based on explanation methods, the ProPublica analysis of the Correctional Offender Management Profiling for Alternative Sanctions (COMPAS) recidivism model \cite{ProPublica, Rudin2019} has revealed that the COMPAS could be racially biased\footnote{Although the analysis might be flawed, see \cite{Rudin2019} for further discussion.}.
    \item \textbf{Mismatched Objectives}: often the objective function of the ML model and the end-goal are mismatched \cite{DoshiVelez2017}. For instance, in a clinical dataset, accurate medical diagnosis may be missing for a substantial group of conservatively treated patients, if the final diagnosis is confirmed only surgically. In this case, the objective function does not provide complete information about performance, and interpreting the model can yield additional insights.
\end{itemize}

From the legal perspective, interpretable and explainable ML is more amenable to the \emph{EU General Data Protection Regulation (GDPR)} \cite{Carvalho2019}\magic{GDPR} that states data subjects' \emph{right to an explanation} of algorithmic decisions and \emph{right to be informed}. It is worth mentioning that the GDPR does not prohibit black box predictive models \cite{Carvalho2019} and that the right to an explanation is not legally binding \cite{Wachter2017}. This, however, as Wachter \emph{et al.} note in \cite{Wachter2017}, does not undermine social and ethical value of providing interpretations and explanations.

\section{Evaluation of Interpretability\label{sec:eval}}

Despite the abundance of methodological research, literature on evaluation metrics for interpretable and explainable ML is scarce \cite{Carvalho2019}. There appears to be no uniform standard for quantitative \emph{in silico} evaluation. Doshi-Velez and Kim \cite{DoshiVelez2017} provide the following classification of evaluation criteria:
\begin{itemize}
    \item \textbf{Application-grounded evaluation} \cite{DoshiVelez2017} requires evaluating a method on an exact task together with human \emph{experts}, representative of the target audience.\magic{application-grounded evaluation} For example, the best way of evaluating an explainable machine-learning-based decision support system for diagnosis would be to ask doctors to perform diagnosis assisted by the system \cite{DoshiVelez2017} and compare their performance to a reasonable baseline. Similar evaluation methods are widely accepted, for example, in the field of \emph{human-computer interaction} \cite{MacDonald2013} and, arguably, if implemented correctly, provide the strongest evidence of success. 
    \item \textbf{Human-grounded evaluation} \cite{DoshiVelez2017} can be seen as a relaxed version of the application-grounded evaluation.\magic{human-grounded evaluation} It requires conducting experiments with human users performing a, possibly, \emph{simplified} task reminiscent of the target application \cite{DoshiVelez2017}. A good example of human-grounded evaluation is the experiment conducted by Ribeiro \emph{et al.} in \cite{Ribeiro2016}. They recruited human subjects on \emph{Amazon Mechanical Turk} and compared their ability to choose the best text classification model based on explanations provided by different techniques \cite{Ribeiro2016}. In this case, human users \emph{were not} experts in the subject area of texts. 
    \item \textbf{Functionally-grounded evaluation} is, arguably, most appropriate for early feasibility studies and is the simplest to implement, since it requires no human subject experiments \cite{DoshiVelez2017}.\magic{functionally-grounded evaluation} These methods use some formal mathematical definition of interpretability as a \emph{proxy measure} \cite{DoshiVelez2017}. For example, Shrikumar \emph{et al.} \cite{Shrikumar2017} performed an experiment evaluating different attribution methods based on the reduction in classification accuracy on the MNIST dataset after \emph{masking} important features identified by an attribution method. 
\end{itemize}
In general, evaluation of interpretability remains largely an open problem. Identifying good proxy metrics and developing rigorous criteria and desiderata for such measures is an important direction for the further research, according to Doshi-Velez and Kim \cite{DoshiVelez2017}.

\begin{figure}[h]
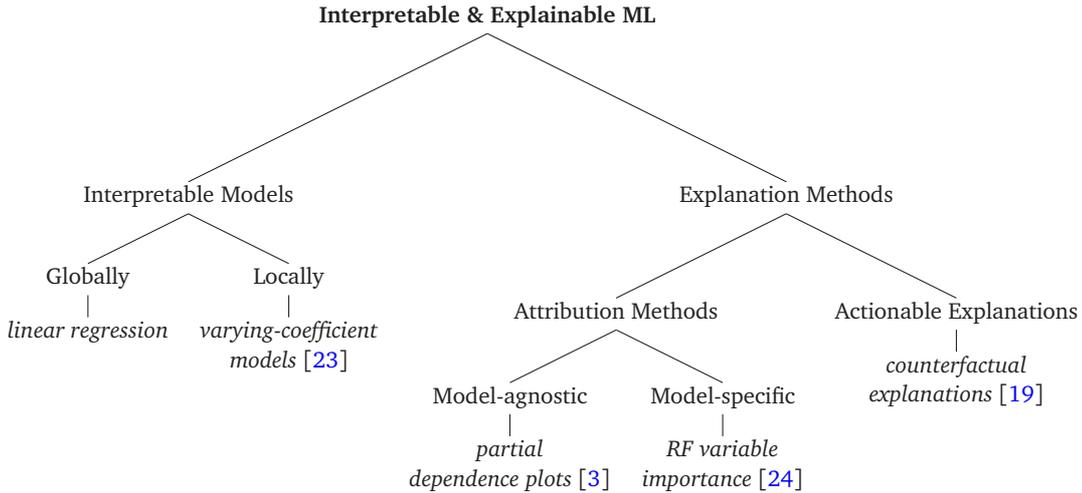

    \begin{center}
        \small
        \Tree[.\textbf{Interpretable \& Explainable ML} [.{Interpretable Models} [.Globally [.\emph{linear regression} ]]
               [.Locally [.{\emph{varying-coefficient}\\ \emph{models} \cite{Hastie1993}} ]]]
          [.{Explanation Methods} [.{Attribution Methods} [.{Model-agnostic} [.{\emph{partial}\\ \emph{dependence plots} \cite{Friedman2001}} ]] [.{Model-specific} [.{\emph{RF variable} \\ \emph{importance} \cite{Breiman2001}} ]]] 
                [.{Actionable Explanations} [.{\emph{counterfactual}\\ \emph{explanations} \cite{Wachter2017}} ]]]]
        \caption{A taxonomy of interpretable and explainable ML techniques, loosely based on the survey by Carvalho \emph{et al.} \cite{Carvalho2019}. Each leaf node represents a simple, well-known example (in \emph{italic}). Note, that this taxonomy is not a \emph{disjoint} partition of the set of all techniques. Moreover, there exist numerous other ways to differentiate between interpretable and explainable ML.}
        \vspace{-0.5cm}
        \label{fig:taxonomy_tree}
    \end{center}
\end{figure}

\section{Examples of Methods\label{sec:methodology}}

In this section, we discuss several state-of-the-art interpretable and explainable ML methods. Figure \ref{fig:taxonomy_tree} provides one possible taxonomy of the field based on the review by Carvalho \emph{et al.} \cite{Carvalho2019}. Every leaf of the tree contains a well-known, simple \emph{example}. For the sake of convenience, we use this taxonomy as a roadmap for our review of the recent literature. Tables~\ref{tab:interpretmethodsummary} and \ref{tab:explmethodsummary} summarise properties of the techniques reviewed in this section. The selection of methods is by no means exhaustive and is meant to illustrate the commonest desirable properties for interpretable models and explanation techniques using concrete instances.

\subsection{Interpretable Models\label{sec:interpretablemodels}}

Interpretable models are usually constrained and structured to reflect physical constraints, monotonicity, additivity, causality, sparsity and/or other desirable properties \cite{Rudin2019, Carvalho2019}. Lipton \cite{Lipton2018} notes that a high-dimensional linear model is not more \emph{intrinsically} interpretable than a compact neural network; on the other hand, a \emph{sparse} linear model is comprehensible and easy to visualise. Two desirable properties, according to \cite{Lipton2018}, are \emph{simulatability} and \emph{decomposability},\magic{simulatability, decomposability} \emph{i.e.} a model must be comprehensible in a limited time and its inputs and parameters should be intuitively meaningful. Below we provide several recent examples of machine learning models that fall into this broad category.

\subsubsection{Falling Rule Lists}

Rule-based classification algorithms have been known for a long time.\magic{rule-based classification} One could argue that these well-established techniques \emph{are} intrinsically interpretable. While single \emph{if-then} rules are indeed comprehensible, \emph{inductive logic programming} \cite{Raedt2010}, for instance, yields an unordered set of conjunctive rules; on the other hand, \emph{decision trees} \cite{Strobl2009} are not \emph{monotonic}\magic{monotonicity} (thus, require additional mental effort) and rely on greedy partition procedures, which only explore a narrow solution subspace. Wang and Rudin \cite{Wang2015} propose a new rule-based \emph{binary} classification model that tries to address mentioned shortcomings -- \emph{falling rule lists} (FLR). The model is motivated by the wide adoption of \emph{risk scores} and \emph{risk stratification} systems in the medical domain.\magic{risk stratification}

A falling rule list is a list of if-then rules s.t. (\emph{i}) during classification, rules have to be applied in the order given by the list, and (\emph{ii}) the probability of the case is monotonically decreasing within the list.\magic{falling rule lists} Wang and Rudin provide an example of an FLR for predicting the risk of malignancy in mammographic masses \cite{Wang2015}. The restricted form of FLRs makes them more comprehensible than decision trees and is natural for practical decision-making in a clinical setting. FLRs are learnt using a Bayesian modelling approach, wherein \emph{monotonicity} and \emph{sparsity} constraints are encoded in the prior distribution. The simulated annealing procedure is used to sample from the posterior distribution and obtain the MAP estimator \cite{Wang2015}. Wang and Rudin argue that, in practice, the sacrifice of performance due to a very restricted form is often not considerable, yet gains in interpretability are substantial.  

Chen and Rudin \cite{Chen2018} further relax the original optimisation problem of learning FLRs by introducing \emph{softly} falling rule lists. As opposed to \cite{Wang2015}, rather than having hard monotonicity constraints, they introduce a non-monotonicity \emph{penalty term}. This formulation is likely better accommodated to noisy real-world datasets, where all sparse solutions might happen to be infeasible. Another interesting extension are \emph{causal} falling rule lists \cite{Wang2015b} that apply FLRs to quantify \emph{treatment effects} in the \emph{potential outcomes framework} \cite{Rubin2005}\magic{treatment effect}.

\subsubsection{Optimised Risk Scores\label{sec:riskscores}}

Another interesting line of research, likewise motivated by medical risk scoring,\magic{risk scores} are supersparse linear integer models (SLIM), introduced by Ustun and Rudin in \cite{Ustun2015}.\magic{supersparse linear models} As opposed to FLRs \cite{Wang2015, Wang2015b, Chen2018} the key focus of which is monotonicity, SLIMs represent \emph{sparse} decision boundaries, \emph{i.e.} relying on a limited number of predictor variables.\magic{sparsity} Moreover, interpretability is facilitated by learning a \emph{linear} scoring function with \emph{integral} coefficients.

Roughly, SLIMs are a special case of the following optimisation problem (see \cite{Ustun2015} for the full formulation):
\begin{equation}
    \min_{\boldsymbol{\beta}}\frac{1}{N}\sum_{i=1}^N\mathbbm{1}_{\left\{y_i\boldsymbol{\beta}^T\matr{x}_i\leq0\right\}}+\lambda_0\left\|\boldsymbol{\beta}\right\|_0+\lambda_1\left\|\boldsymbol{\beta}\right\|_1,\text{ s.t. }\boldsymbol{\beta}\in\mathcal{B},
    \label{eqn:slim}
\end{equation}
where $\boldsymbol{\beta}\in\mathcal{B}$ is an integer-valued coefficient vector with $\mathcal{B}=\left\{L,L+1,...,U-1,U\right\}^p$, $L<U$; $(\matr{x}_i,y_i)$ are labelled data points; and $N$ is the size of the training set. The integer linear program (ILP) defined in Equation~\ref{eqn:slim} enjoys the advantages of \emph{directly} minimising the 0-1 loss and the $\ell_0$ penalty instead of convex surrogate measures, commonly adopted in statistics and machine learning literature, \emph{cf.} \cite{Zou2005}.\magic{$\ell_0$ penalty} Given recent improvements in ILP solvers, such problems have become tractable even for datasets with 1000s of data points: Ustun and Rudin \cite{Ustun2015} report a running time of under 10 minutes for $N=1922$.

In addition to the theoretical guarantees, the ILP formulation above has a benefit of easily incorporating and enforcing additional constraints \cite{Ustun2015}, beyond integrality and sparsity. For example, introducing desirable `either-or' or `if-then' conditions on features or preferences for (not) using certain predictors. Ustun and Rudin also introduce a range of extensions/generalisations of SLIMs. Particularly interesting are \emph{personalised} SLIMs -- a soft version with scoring rules differing for individual data points. As opposed to SLIMs that are globally interpretable, personalised SLIMs are only locally interpretable. The authors also present rule-based adaptations \cite{Ustun2015}. Further algorithmic improvements to this approach are made by Ustun and Rudin in \cite{Ustun2017}.

\subsubsection{Generalised Additive Models\label{sec:gams}}

As mentioned before, decomposability is a desirable property of interpretable ML models \cite{Lipton2018}. A noteworthy class of decomposable functions are \emph{additively separable} functions \cite{Segal1994}.\magic{additivity} We say that function $f(x_1,x_2,...,x_p)$ is additively separable if we can rewrite it in the form $f(x_1,x_2,...,x_p)=\sum_{j=1}^pu_j(x_j)$ \cite{Segal1994}. Hastie and Tibshirani \cite{Hastie1986} introduce the class of \emph{generalised additive models} (GAM) that rely on this additivity property. In particular, for $p$ predictor variables, a GAM is given by 
\begin{equation}
    g(y)=\sum_{j=1}^ps_j(x_j),
    \label{eqn:gams}
\end{equation} 
where $g(\cdot)$ is a \emph{link function}, $s_j(\cdot)$ are smooth functions \cite{Hastie1986}, which we will refer to as \emph{shape functions} \cite{Lou2012}.\magic{generalised additive models} This class can be seen as an extension of the linear model that preserves the additivity, but allows introducing nonlinearities in individual predictors by choosing appropriate shape functions.

Lou \emph{et al.} \cite{Lou2012} argue in favour of the intelligibility of GAMs: this class of models ignores \emph{interactions} between predictor variables, therefore, influence of each predictor is easily comprehensible and can be visualised. They conduct extensive experimental comparison among different methods for fitting GAMs and choices of $s_j(\cdot)$. They consider \emph{least squares}, \emph{gradient boosting} (with slight adjustments to the original procedure in \cite{Friedman2001}), and \emph{backfitting}. In addition to the standard use of \emph{spline-based} shape functions (referred to as \emph{scatterplot smoothers} in \cite{Hastie1986}), they consider \emph{single}, \emph{bagged}, \emph{boosted}, and \emph{boosted bagged decision trees} \cite{Lou2012}.

Building on \cite{Lou2012}, Caruana \emph{et al.} \cite{Caruana2015} propose a simple, yet more performant extension by including two-way interaction terms, referred to as GA\textsuperscript{2}M (\emph{cf.} Equation~\ref{eqn:gams}):
\begin{equation}
    g(y)=\sum_{j=1}^ps_j(x_j)+\sum_{j=1}^p\sum_{\substack{k=1 \\ k\neq j}}^ps_{jk}(x_j, x_k),
    \label{eqn:ga2ms}
\end{equation} 
where $s_{jk}(x_j,x_k)$ are pairwise interaction terms. Caruana \emph{et al.} argue that pairwise interactions are still intelligible, since they can be visualised using heat maps \cite{Caruana2015}.\magic{interactions} The authors provide a pneumonia risk prediction case study showing that GA\textsuperscript{2}Ms can perform on par with uninterpretable techniques in tabular datasets, while being more amenable to \emph{verification} and \emph{debugging}.

Another interesting extension of additive regression modelling are sparse additive models (SpAM), proposed by Ravikumar \textit{et al.} in \cite{Ravikumar2007}. This model class combines ideas of Hastie and Tibshirani \cite{Hastie1986} with sparse linear modelling for high-dimensional regression problems, \textit{e.g}. \cite{Zou2005}. In addition to shape functions $s_j(\cdot)$, Ravikumar \textit{et al.} \cite{Ravikumar2007} introduce a weight vector $\boldsymbol{\beta}\in\mathbb{R}^p$ multiplied with outputs of $s_j(\cdot)$ and penalise its norm in the loss function. In this way, SpAMs rely on a \textit{sparse} subset of features and still preserve the \textit{additive} structure of the classical GAMs.

\subsubsection{Sparse-input Neural Networks\label{sec:spinn}}

In many high-dimensional problems, \emph{e.g.} in genomic data analysis \cite{Lucas2006}, social network modelling \cite{Ravazzi2018} etc., sparsity is an important inductive bias, which allows deducing parsimonious interpretations. We have already mentioned sparsity as a desirable property when describing supersparse linear integer models in Section~\ref{sec:riskscores}. Recently, there have been renewed efforts in leveraging sparsity-inducing regularisation to understand and control the behaviour of neural networks models \cite{Feng2017, Lu2018, Tank2018, Khanna2020}. 

Feng and Simon \cite{Feng2017} provide a thorough analysis, both theoretical and empirical, of \emph{sparse-input neural networks} (SPINNs) in the context of $p\gg N$ problems \cite{Hastie2009}.\magic{sparse-input neural networks} SPINNs  are trained by solving the following optimisation problem:
\begin{equation}
    \min_{\boldsymbol{\theta}}\frac{1}{N}\sum_{i=1}^N\ell\left(y_i,f_{\boldsymbol{\theta}}(\matr{x}_i)\right)+\lambda_0\sum_{a=2}^L\left\|\matr{W}_a\right\|^2_2 + \lambda\sum_{j=1}^p\Omega_{\alpha}\left(\matr{W}_{1,\cdot j}\right),
    \label{eqn:spinn}
\end{equation} 
where $\boldsymbol{\theta}=\left\{\matr{W}_a\right\}_{a=1}^L$ are weight matrices; $\matr{W}_{1,\cdot j}$ refers to the $j$-th column of the input layer weight matrix; and $\Omega_{\alpha}\left(\boldsymbol{\beta}\right)=(1-\alpha)\|\boldsymbol{\beta}\|_1+\alpha\|\boldsymbol{\beta}\|_2$ with $\alpha\in(0,1)$ is the \emph{sparse group Lasso} penalty.\magic{Lasso penalty} Here, parameter $\alpha$ controls the trade-off between the element-wise Lasso and the group Lasso penalties. Feng and Simon \cite{Feng2017} prove probabilistic, finite-sample generalisation guarantees for this model class and demonstrate performance gains empirically for high-dimensional data with higher-order interactions, when compared to other nonparametric models. Tank \emph{et al.} \cite{Tank2018} and Khanna and Tan \cite{Khanna2020} leverage similar penalties in the context of autoregressive time series modelling and Granger-causal inference (see Section~\ref{sec:tsa}).

\subsubsection{Knockoff Features}

The ultimate goal of SPINNs \cite{Feng2017} is ``\emph{deep}'' feature selection.\magic{feature selection} Lu \emph{et al.} \cite{Lu2018} propose another solution: \emph{deep feature selection using paired-input nonlinear knockoffs} (DeepPINK) which leverages \emph{knockoff filters} to facilitate interpretability and sparsity in deep neural networks at the input level.\magic{knockoff filters} An important advantage of this technique over SPINNs \cite{Feng2017} is that it provides a \emph{false discovery rate} (FDR) control \cite{Benjamini1995} when selecting significant features.\magic{false discovery rate}

Knockoff filters were originally proposed by Barber and Candès in \cite{Barber2015}. In brief, knockoff filters are a variable selection procedure that controls the FDR exactly in a linear model in finite sample settings, whenever there are at least as many observations as predictor variables. The key idea is to construct knockoff features mimicking the dependency structure of the original features; augment the original dataset with knockoffs; and compare statistics for each original feature and its knockoff, \emph{e.g.} the absolute value of the coefficient. In this way, variables with genuine signals can be identified while controlling for the FDR. Candès \emph{et al.} \cite{Candes2018} propose `\emph{model-$X$}' knockoffs which rely on the assumption that the joint distribution of covariates is known, without assuming \emph{anything} about the distribution of the response variable conditional on predictors. The key limitation of \cite{Candes2018} is that the generation of knockoff features is based on a known multivariate Gaussian distribution. Jordon \emph{et al.} \cite{Jordon2018} alleviate this issue by introducing the \emph{KnockoffGAN} -- a generative adversarial network (GAN) \cite{Goodfellow2014} for knockoff generation capable of learning complex dependency structures. Along similar lines, following the model-$X$ framework, Romano \emph{et al.} \cite{Romano2019} propose constructing knockoff features with deep generative models utilising the maximum
mean discrepancy (MMD) \cite{Li2015}.

\subsubsection{Varying-coefficient Models\label{sec:vcm}}

While GAMs \cite{Hastie1986} generalise the linear model by allowing for nonlinearities in predictor variables (see Section~\ref{sec:gams}), \emph{varying-coefficient models} (VCM),\magic{varying-coefficient models} proposed by Hastie and Tibshirani in \cite{Hastie1993}, offer a different sort of generalisation. In a VCM, covariate coefficients vary smoothly with so called `\textit{effect modifiers}' --- variables $r_1,r_2,...,r_p$:
\begin{equation}
    g(y)=\beta_0+\sum_{j=1}^px_j\beta_j(r_j),
    \label{eqn:vcm}
\end{equation} 
wherein $\beta_j(\cdot)$ are unspecified smooth functions corresponding to the varying coefficients of predictor variables. The choice of variables $r_1,...,r_p$ depends on a particular application; for example, $r_j$ may coincide with predictors $x_j$ or correspond to some exogenous variables. In dynamical systems, \emph{time} can be a single effect modifier, producing \emph{time-varying} coefficients. 

Clearly, VCMs are only locally interpretable, since coefficients vary across data points. Nevertheless, this trade-off may be desirable in pursuit of a more flexible model. Some of the interpretable ML models discussed further \cite{AlShedivat2017, AlvarezMelis2018} bear a striking resemblance to VCMs: they generalise this classical framework to unstructured data sources by parameterising $\beta_j(\cdot)$ with neural networks. 

\subsubsection{Contextual Explanation Networks\label{sec:cen}}

The `\emph{credit assignment problem}' in multilayer neural networks is difficult \cite{Guerguiev2017}, largely limiting the interpretability of this family of models.\magic{credit assignment problem} Quantification and explanation of contributions of individual inputs to the resulting predictions is not straightforward, due to entangled interactions in downstream layers \cite{Guerguiev2017, Tank2018}. Nevertheless, there has been a substantial effort to explain neural network models \emph{post hoc}, \emph{i.e.} after training: \cite{Bach2015}, \cite{Shrikumar2017}, and \cite{Selvaraju2019} are just a few examples (see Section~\ref{sec:attrib} for an overview of similar techniques).

To this end, Al-Shedivat \emph{et al.} \cite{AlShedivat2017} propose \emph{contextual explanation networks} (CEN) -- a class of neural network architectures that jointly predict and explain their predictions without requiring additional model introspection.\magic{contextual explanation networks} CENs are defined as deep probabilistic models for learning $\mathbf{P}_{\matr{w}}\left(\matr{Y}\:|\:\matr{x},\matr{c}\right)$, parameterised by $\matr{w}$, where $c\in\mathcal{C}$ are \emph{context} variables, $\matr{x}\in\mathcal{X}$ are \emph{attributes}, and $\matr{y}\in\mathcal{Y}$ are \emph{targets}, to be predicted from $\matr{x}$ and $\matr{c}$. Given $\matr{x}$, $\matr{y}$, and $\matr{c}$, the probabilistic model defined in \cite{AlShedivat2017} is of the following form:
\begin{equation}
    \begin{split}
        \matr{y}&\sim\mathbf{P}\left(\matr{Y}\:|\:\matr{x},\boldsymbol{\theta}\right),\\
        \boldsymbol{\theta}&\sim\mathbf{P}_{\matr{w}}\left(\boldsymbol{\theta}\:|\:\matr{c}\right),\\
        \mathbf{P}_{\matr{w}}\left(\matr{Y}\:|\:\matr{x},\matr{c}\right)&=\int\mathbf{P}\left(\matr{Y}\:|\:\matr{x},\boldsymbol{\theta}\right)\mathbf{P}_{\matr{w}}\left(\boldsymbol{\theta}\:|\:\matr{c}\right)d\boldsymbol{\theta},
    \end{split}
    \label{eqn:cen}
\end{equation}
where $\mathbf{P}\left(\matr{Y}\:|\:\matr{x},\boldsymbol{\theta}\right)$ is a predictor parameterised by $\boldsymbol{\theta}$, explicitly relating attributes to targets. $\boldsymbol{\theta}$ can be seen as an \emph{explanation} that is specific to the context given by $\matr{c}$. In practice, $\mathbf{P}_{\matr{w}}\left(\boldsymbol{\theta}\:|\:\matr{c}\right)$ is replaced with an encoder neural network and the predictive distribution $\mathbf{P}\left(\matr{Y}\:|\:\matr{x},\boldsymbol{\theta}\right)$ is specified by an interpretable function, \emph{e.g.} the linear model $f_{\boldsymbol{\theta}}(\matr{x})=\textrm{softmax}\left(\boldsymbol{\theta}^\top\matr{x}\right)$.

The relationship between VCMs \cite{Hastie1993} and CENs \cite{AlShedivat2017} is clear: contextual variables $\matr{c}$ can be seen as effect modifiers for attributes $\matr{x}$ (\emph{cf.} Equation~\ref{eqn:vcm}). The key contribution of CENs is to cast the classical generalisation of linear models into a probabilistic framework while parameterising coefficients with neural networks. Al-Shedivat \emph{et al.} demonstrate the efficacy of their approach on classification and survival analysis \cite{Altman2020} tasks and show that CENs are still interpretable in datasets with noisy features where \textit{post hoc} explanation techniques can often be inconsistent and misleading.
\newpage

\subsubsection{Self-explaining Neural Networks\label{sec:senn}}

Another interesting class of functions was introduced by Alvarez-Melis and Jaakkola in \cite{AlvarezMelis2018}. Similar to Al-Shedivat \emph{et al.} \cite{AlShedivat2017}, they propose an \emph{intrinsically} interpretable neural network model that allows disentangling contributions of individual predictor variables or basis concepts. Self-explaining neural networks (SENN) \cite{AlvarezMelis2018} are motivated by \emph{explicitness}, \emph{faithfulness}, and \emph{stability} properties -- three desiderata for interpretability.\magic{self-explaining neural networks} SENNs act like a simple model locally, but can be highly complex and nonlinear globally:
\begin{equation}
    f(\matr{x})=\boldsymbol{\theta}(\matr{x})^\top\matr{x},
    \label{eqn:gencoeffs}
\end{equation}
where $\matr{x}\in\mathbb{R}^p$ are predictors; and $\boldsymbol{\theta}(\cdot)$ is a neural network with $p$ outputs. $\boldsymbol{\theta}(\matr{x})$ are \emph{generalised coefficients} for data point $\matr{x}$. Without further restrictions the model in Equation~\ref{eqn:gencoeffs} is not more interpretable than a simple multilayer neural network. We want the model to be locally linear: it needs to hold that $\nabla_{\matr{x}}f(\matr{x})\approx\boldsymbol{\theta}(\matr{x}_0)$, for all $\matr{x}$ in the neighbourhood of $\matr{x}_0$. Under this constraint, individual components of $\boldsymbol{\theta}(\matr{x})$ are interpretable and adaptive coefficients.

Further extensions of the model given by Equation~\ref{eqn:gencoeffs} are possible. For example, we can introduce \emph{basis concepts} $\boldsymbol{h}(\matr{x}):\mathbb{R}^p\rightarrow\mathbb{R}^k$ and use them instead of the raw predictors: $f(\matr{x})=\boldsymbol{\theta}(\matr{x})^\top \boldsymbol{h}(\matr{x})$. Furthermore, a general link function $g(\cdot)$ can be used, resulting in
\begin{equation}
    f(\matr{x})=g\left(\theta(\matr{x})_1h(\matr{x})_1,...,\theta(\matr{x})_kh(\matr{x})_k\right),
    \label{eqn:linkbasis_senn}
\end{equation} 
where $z_j=\theta\left(\matr{x}\right)_jh(\matr{x})_j$ is the \emph{influence score} of the $j$-th concept.\magic{influence score}

In practice, a SENN given by Equation~\ref{eqn:linkbasis_senn} is trained by minimising the following \emph{gradient-regularised} loss function \cite{AlvarezMelis2018}, which balances performance with interpretability:
\begin{equation}
    \mathcal{L}_y(f(\matr{x}),y)+\lambda\mathcal{L}_{\boldsymbol{\theta}}\left(f(\matr{x})\right),
    \label{eqn:sennloss}
\end{equation}
where $\mathcal{L}_y(f(\matr{x}),y)$ is a loss term for the ground classification or regression task, \emph{e.g.} the mean squared error or the cross entropy; $\lambda>0$ is a regularisation parameter; and $\mathcal{L}_{\boldsymbol{\theta}}(f(\matr{x}))$ is the gradient penalty:\magic{gradient penalty}
\begin{equation}
    \mathcal{L}_{\boldsymbol{\theta}}\left(f(\matr{x})\right)=\left\|\nabla_{\matr{x}}f(\matr{x})-\boldsymbol{\theta}(\matr{x})^\top \matr{J}^{\boldsymbol{h}}_{\matr{x}}(\matr{x})\right\|_2,
    \label{eqn:gradpenalty}
\end{equation}
where $\matr{J}^{\boldsymbol{h}}_{\matr{x}}$ is the Jacobian of $\boldsymbol{h}(\cdot)$ w.r.t. $\matr{x}$.

Following are important restricting assumptions, crucial for enforcing interpretability:
\begin{enumerate}
    \item $g(\cdot)$ (see Equation~\ref{eqn:linkbasis_senn}) is \emph{monotonic} and \emph{additively separable} in its arguments (see Section~\ref{sec:gams});
    \item $\frac{\partial g}{\partial z_i}>0$ with $z_i=\theta(\matr{x})_ih(\matr{x})_i$, for all $i$;
    \item $\boldsymbol{\theta}(\cdot)$ is \emph{locally difference-bounded} by $\boldsymbol{h}(\cdot)$, \emph{i.e.} for every $\matr{x}_0$, there exist $\delta>0$ and $L\in\mathbb{R}$ s.t. if $\left\|\matr{x}-\matr{x}_0\right\|_2<\delta$, then $\left\|\boldsymbol{\theta}(\matr{x})-\boldsymbol{\theta}(\matr{x}_0)\right\|_2\leq L\left\|\boldsymbol{h}(\matr{x})-\boldsymbol{h}(\matr{x}_0)\right\|_2$;
    \item $\left\{h(\matr{x})_i\right\}_{i=1}^k$ are interpretable representations\footnote{Alvarez-Melis and Jaakkola \cite{AlvarezMelis2018} emphasise three guiding criteria for choosing \emph{interpretable} basis concepts: (\emph{i}) \emph{fidelity}: representations should contain relevant context information; (\emph{ii}) \emph{diversity}: concepts used to represent inputs should be few and non-overlapping; (\emph{iii}) \emph{grounding}: concepts should be immediately understandable to a human. Moreover, they explain how such representations can be learnt using autoencoders in an end-to-end manner in conjunction with the SENN model.} of $\matr{x}$;
    \item $k$ is small.
\end{enumerate}
The class of functions described by assumptions 1-5 is fairly flexible, \emph{e.g.} \emph{generalised linear models} \cite{Nelder1972} and the nearest neighbour classifier satisfy these assumptions \cite{AlvarezMelis2018}.\magic{generalised linear models} The key advantage of SENNs stems from the richness of complex neural network architectures used for $\boldsymbol{\theta}(\cdot)$ and $\boldsymbol{h}(\cdot)$.

Quinn \emph{et al.} \cite{Quinn2020} further extend SENNs to \emph{compositional data analysis} (CoDA), such as high-throughput sequencing data. In CoDA, the log-ratio transform is often applied with some reference value in the denominator.\magic{compositional data} Quinn \emph{et al.} validate their framework on 25 microbiome datasets, demonstrating benefits of nonlinearity and \emph{personalised} interpretability \cite{Quinn2020} achieved by SENNs.

Like CENs \cite{AlShedivat2017}, SENNs \cite{AlvarezMelis2018} are clearly related to varying-coefficient models \cite{Hastie1993} (see Section~\ref{sec:vcm}). The key difference being that in SENNs, regressors themselves act as effect modifiers and that the framework is augmented with interpretable basis concepts, defined on top of the raw inputs. Certainly, all three model classes discussed above \cite{Hastie1993, AlShedivat2017, AlvarezMelis2018} hold a promise of local interpretability while providing room for predictively powerful models.

\subsubsection{Attentive Mixtures of Experts}

In natural language processing, the \emph{attention mechanism} has become a powerful tool for exploring relationships between inputs and outputs of deep learning models \cite{Vaswani2017}, utilised both for interpretability and performance.\magic{attention mechanism} Nevertheless, several works critique the na\"{i}ve use of attention for model interpretation \cite{Jain2019, Serrano2019}, showing empirically that it is often uncorrelated with gradient information and other natural feature importance measures. Schwab \emph{et al.} \cite{Schwab2019} propose a \emph{mixture of experts} model with attentive gates for learning feature importance values alongside with making predictions, introducing an auxiliary objective to mitigate the aforementioned shortcomings of the the na\"{i}ve attention.\magic{mixture of experts}

Let us consider predicting a target variable $y$ based on features $\matr{x}\in\mathbb{R}^p$. The \emph{attentive mixture of experts} (AME) model, proposed by Schwab \emph{et al.} \cite{Schwab2019}, consists of several \emph{connected} experts.\magic{attentive mixture of experts} The model is given by the following equations:
\begin{equation}
    \begin{split}
        f(\matr{x})&=\sum_{j=1}^p\underbrace{G_j\left(\matr{h}_{all}\right)}_{a_j}\underbrace{E_j(x_j)}_{c_j},\\
        \matr{h}_{all}&=\left[h_1, c_1, h_2, c_2, \dots, h_p, c_p\right],\\
        a_j&=\frac{\exp\left(\matr{u}_j^\top\matr{u}_{s,j}\right)}{\sum_{k=1}^p\exp\left(\matr{u}_k^\top\matr{u}_{s,k}\right)}, \quad \textrm{for $1\leq j\leq p$},\\
        \matr{u}_j&=\sigma\left(\matr{W}_j\matr{h}_{all}+b_j\right), \quad \textrm{for $1\leq j\leq p$},
    \end{split}
    \label{eqn:ame}
\end{equation}
where $c_j=E\left(x_j\right)$ is the output of the $j$-th expert; $a_j$ is the output of the attentive gating network $G_j(\cdot)$ and corresponds to the importance of the $j$-th feature; $h_j$ denotes a hidden state from the $j$-th expert; $\matr{u}_j$ is a projected representation of $\matr{h}_{all}$; $\matr{u}_{s,j}$ is a per-expert learnable context vector; and $\sigma(\cdot)$ is an activation function.

The AME is trained end-to-end by minimising a loss function augmented with an auxiliary objective. The auxiliary objective encourages the importance score $a_j$ to reflect the decrease in error associated with the contribution of the $j$-th expert, similar in spirit to the definition of the RF variable importance \cite{Breiman2001}:
\begin{equation}
    \Delta\varepsilon_{\matr{x},j}=\varepsilon_{\matr{x}\setminus\{j\}}-\varepsilon_{\matr{x}},\quad\textrm{for $1\leq j\leq p$,}
    \label{eqn:errdelta}
\end{equation}
where $\varepsilon_{\matr{x}\setminus\{j\}}$ and $\varepsilon_{\matr{x}}$ denote the prediction error of the model without the $j$-th feature and of the full model, respectively. Desired attribution scores $\Delta\varepsilon_{\matr{x},j}$ are normalised to produce
\begin{equation}
    \omega_{\matr{x},j}=\frac{\Delta\varepsilon_{\matr{x},j}}{\sum_{k=1}^p\Delta\varepsilon_{\matr{x},k}},\quad\textrm{for $1\leq j\leq p$.}
    \label{eqn:errdeltanorm}
\end{equation}
The auxiliary loss term is then given by the discrepancy between the normalised desired attribution scores and the outputs of attentive gating networks:
\begin{equation}
    \mathcal{L}_{aux}=\frac{1}{n}\sum_{i=1}^nD\left(\boldsymbol{\omega}_{\matr{x}_i},\boldsymbol{a}_{\matr{x}_i}\right),
    \label{eqn:ameaux}
\end{equation}
where $D(\cdot,\cdot)$ is the discrepancy measure, \emph{e.g.} the Kullback--Leibler divergence; $\boldsymbol{\omega}_{\matr{x}_i}$ and $\boldsymbol{a}_{\matr{x}_i}$ are normalised desired attribution scores and the outputs of attentive gating networks, respectively, for the $i$-th data point.

The attentive mixture of experts \cite{Schwab2019} successfully overcomes limitations of the na\"{i}vely trained attention mechanism by introducing regularisation, which forces learning importance scores reflective of the prediction error. Next to CENs \cite{AlShedivat2017} and SENNs \cite{AlvarezMelis2018}, AMEs are yet another class of locally interpretable neural network architectures which produce individualised explanations for each data point at prediction time. The relationship between predictions and explanations is, arguably, more opaque within AMEs than within the previously described model classes.
\newpage

\subsubsection{Symbolic Regression\label{sec:symreg}}

\emph{Symbolic regression} is a problem of inferring an \emph{analytic} form for an unknown function which can only be queried \cite{AmirHaeri2017, Udrescu2020}.\magic{symbolic regression} While neural networks do provide analytic representations, symbolic regression usually seeks \emph{parsimonious} equations, \emph{e.g.} making further restrictions to \emph{polynomial} forms. Thus, symbolic regression can be leveraged to learn interpretable functional relationships \cite{Jin2019} from raw data. For example, when predicting target $y$ based on features $\matr{x}\in\mathbb{R}^p$, a mathematical expression $g(\matr{x})=x_1+2\cos(x_2)+\exp(x_3)+0.1$ could be a candidate solution for symbolic regression \cite{Jin2019}. The optimisation problem can be formalised as follows:
\begin{equation}
    \min_{g\in\mathfrak{G}}\frac{1}{n}\sum_{i=1}^n\ell\left(g\left(\matr{x}_i\right), y_i\right),
    \label{eqn:symreg}
\end{equation}
where $\mathfrak{G}$ is a set of \emph{succinct} mathematical expressions and $\ell(\cdot,\cdot)$ is the loss function, \emph{e.g.} if $y$ is continuously-valued, a possible choice is $\ell\left(g\left(\matr{x}_i\right), y_i\right)=\left(g\left(\matr{x}_i\right)-y_i\right)^2$.
Some of the traditional approaches to symbolic regression include genetic programming \cite{AmirHaeri2017} and simulated annealing \cite{Stinstra2007}. In Section~\ref{sec:symmeta}, we discuss how symbolic modelling can be leveraged to explain black box models \emph{post hoc} by approximating \emph{e.g.} a neural network with a symbolic \emph{metamodel} \cite{Alaa2019}. 

\subsubsection{Time Series Analysis\label{sec:tsa}}

Interpretable models can be instrumental in understanding and exploring \emph{relationships} in multivariate time course data.\magic{time series} Inference of time-varying relationships has been addressed before with a variety of techniques which usually rely on sparse autoregressive models. An iconic example is the work by Song \emph{et al.} \cite{Song2009} who propose \emph{time-varying dynamic Bayesian networks} (TV-DBN) based on autoregression with kernelised weights and $\ell_1$ regularisation. A similar approach is described by Kolar \emph{et al.} in \cite{Kolar2010} where the authors propose using $\ell_1$-regularised temporally-smoothed logistic regression for estimating \emph{time-varying} dependency networks.  

Both of the approaches \cite{Song2009, Kolar2010} are closely related to the inference of \emph{Granger causality} (GC) \cite{Granger1969, Luetkepohl2007},\magic{Granger causality} the key difference being that GC relationships are \emph{directed} and rely on the \emph{temporal precedence} of the cause before its effect. As mentioned before, one of the desiderata for interpretable and explainable ML is causality (see Section~\ref{sec:intro}). GC has become a popular tool for exploratory analysis in economics \cite{Appiah2018}, neuroscience \cite{Roebroeck2005}, and other domains which require understanding multivariate dynamic stochastic systems. The latest approaches to inferring Granger-causal relationships leverage neural networks \cite{Tank2018, Nauta2019, Khanna2020, Wu2020, Loewe2020} and are often based on interpretable regularised autoregressive models \cite{Tank2018, Khanna2020, Wu2020}, reminiscent of the \emph{Lasso Granger} method \cite{Arnold2007}.\magic{Lasso Granger} 

In general, the problem of \emph{relational learning} \cite{Kipf2018} from time series data is an area of ongoing research.\magic{relational learning} Extensions and adaptations of the existing approaches to different scenarios, such as replicate- and time-varying dependency structures, data scarcity, and high dimensionality, can be of interest to both the ML community and application domain areas. For instance, in the medical domain, inferring relationships from longitudinal data can aid in discovering and understanding \emph{sequential} events in \emph{patient disease trajectories} \cite{Paik2019} and exploring influence of interventions \cite{Bigelow2005}.\magic{patient trajectories}

\subsubsection{Interpretable Representation Learning}

Sometimes it might be desirable to learn low-dimensional \emph{embeddings}, or \emph{representations}, in an unsupervised, weakly- or semi-supervised manner rather than explore relationships between \emph{raw} variables, as most methods discussed in the previous sections do.\magic{representation learning} Representations can be useful in several (usually unknown at the time of representation learning) \emph{downstream applications}. A desirable property that is targeted by some techniques in the representation learning domain is interpretability, which is usually attained by enforcing some constraints on representations. 

For example, InfoGAN \cite{Chen2016}, an information-theoretic extension of GANs, proposed by Chen \emph{et al.} has been empirically shown to be able to learn \emph{disentangled} representations,\magic{disentanglement} \emph{i.e.} representations wherein separate sets of dimensions are uniquely correlated with salient semantically meaningful features of data. Similar results can be achieved with variational autoencoders (VAE), in particular, $\beta$-VAE \cite{Higgins2016} introduces statistical independence constraints on embedding dimensions via a factorising prior distribution. Such representations, for example, allow for a controlled generation of synthetic data. In theory, learning disentangled representations in a completely unsupervised manner without introducing inductive biases has been shown to be fundamentally impossible \cite{Locatello2019}. The latter result, nevertheless, does not diminish the opportunity to learn disentangled representations by injecting inductive biases and implicit or explicit forms of supervision \cite{Locatello2019}.

One line of research in this direction focuses on learning \emph{structured} representations of \emph{sequential} data \cite{Tschannen2018}. For instance, Li and Mandt \cite{Li2018} propose a sequential VAE architecture that disentangles representations into \emph{static} and \emph{dynamic} parts, providing partial control over \emph{content} and \emph{dynamics} of generated sequences.\magic{content \& dynamics} In \cite{Hsu2017}, Hsu \emph{et al.} explore a somewhat similar problem and approach. There are several other works targeting the disentanglement of static and dynamic components in representations \cite{Tschannen2018}. On the other hand, Miladinović \emph{et al.} \cite{Miladinovic2019} propose \emph{disentangled state space models} and consider a slightly different setting wherein sequences are observed in different \emph{environments}, or \emph{domains}. Their aim is to discern domain-specific and domain-invariant factors of variation.

We refer the interested reader to the paper by Tschannen \emph{et al.} \cite{Tschannen2018} for an exhaustive (slightly outdated) overview of advancements in autoencoder-based representation learning. Many of the approaches discussed in \cite{Tschannen2018} strive towards some form of interpretability in learnt representations, although often do not state that explicitly.

\bigskip

Interpretable representation learning certainly goes beyond disentanglement. For example, Manduchi \emph{et al.} \cite{Manduchi2019} propose \emph{deep probabilistic self-organising maps} (DPSOM) -- a VAE-based neural network architecture for probabilistic clustering.\magic{clustering}\magic{self-organising maps} Self-organising maps (SOM), originally proposed by Kohonen in\cite{Kohonen1982}, provide a natural and interpretable discrete visualisation of high-dimensional data. Thus, DPSOM \cite{Manduchi2019} learns \textit{rich} representations for probabilistic clustering which can be presented in a human-comprehensible form. Manduchi \emph{et al.} showcase the DPSOM by visualising patient state trajectories throughout time.

Another noteworthy example of interpretable representation learning are \emph{concept bottleneck models} \cite{Kumar2009, Lampert2009} re-explored by Koh \emph{et al.} in \cite{Koh2020}.\magic{concept bottleneck models} As opposed to previous techniques, concept bottleneck models are completely supervised. For predicting target $y$ based on features $\matr{x}\in\mathbb{R}^p$, a bottleneck model is given by $f\left(g\left(\matr{x}\right)\right)$, where $g:\:\mathbb{R}^p\to\mathbb{R}^k$ and $f:\:\mathbb{R}^k\to\mathbb{R}$. In \textit{concept} bottlenecks, $f(\cdot)$ relies completely on concepts $\matr{c}=g\left(\matr{x}\right)$, and $g(\cdot)$ is learnt in a supervised manner using labelled interpretable concepts. Koh \emph{et al.} explore a range of strategies for training such models \cite{Koh2020}. In practice, $f(\cdot)$ and $g(\cdot)$ are parameterised by neural networks. An important advantage of concept bottlenecks is that at prediction time, an expert, \emph{e.g.} a medical doctor, can intervene on incorrectly inferred concepts. Arguably, applicability of concept bottleneck models is limited to areas and tasks where vast domain knowledge is available and where instances can be labelled by experts.

\medskip

To summarise, the problem of interpretability in representation and, more broadly, \emph{unsupervised learning} is still under-explored, despite a growing body of research \cite{Adel2018, Manduchi2019, Koh2020, Dasgupta2020}.\magic{unsupervised learning} As can be seen from previous sections, many techniques focus exclusively on classification and regression tasks, while deep clustering, generative modelling, and representation learning have attracted comparatively less attention. With the emergence of new socially consequential application domains for these problems, a need for interpretable techniques is becoming apparent. 

\subsection{Explanation Methods\label{sec:explainable}}

We now turn towards a completely different family of methods. According to Rudin \cite{Rudin2019}, explainable ML focuses on providing explanations for existing black box models by the means of training another \emph{surrogate} model \emph{post hoc}. Explanations can take various forms: textual, visual, symbolic; even a data point from the training set or a synthetic data point can serve as an explanation \cite{Lipton2018}. An explanation can be \emph{global}, \emph{i.e.} characterise the whole dataset, \emph{e.g.} random forest variable importance \cite{Breiman2001}, or \emph{local},\magic{locality} \emph{i.e.} explain individual classification or regression outcomes \cite{Carvalho2019}, \emph{e.g.} generalised coefficients $\boldsymbol{\theta}(\matr{x})$ in SENNs (see Section~\ref{sec:senn} and Equations~\ref{eqn:gencoeffs} and \ref{eqn:linkbasis_senn}) can be seen as a local explanation for data point $\matr{x}$. It can be \emph{model-specific}, \emph{i.e.} capable of explaining only a restricted class of models, or \emph{model-agnostic},\magic{model agnosticism} \emph{i.e.} applicable to an arbitrary model \cite{Carvalho2019}.  Carvalho \emph{et al.} summarise following desirable properties of an explanation technique \cite{Carvalho2019}:\magic{desiderata for explainability}
\begin{itemize}
    \item \textbf{Faithfulness}: an explanation model should be faithful w.r.t. the black box model it tries to explain, \emph{i.e.} predictions produced by the two models should agree as much as possible.
    \item \textbf{Consistency \& Stability}: explanations for different models tackling the same task should be consistent, and explanations for similar data points should be similar as well. These properties might be difficult to attain, especially, considering the \emph{Rashomon effect}\footnote{Named after the A. Kurosawa's 1950 film ``\emph{Rashōmon}''.} \cite{Semenova2019, Roth2002} which occurs when multiple equally feasible explanations exist for one phenomenon.\magic{Rashomon effect}
    \item \textbf{Comprehensibility}: an explainee should be able to easily comprehend explanations. This property is related to simulatability posited by Lipton in \cite{Lipton2018} for interpretable models (see Section~\ref{sec:interpretablemodels}).
    \item \textbf{Certainty \& Novelty}: explanations should convey (un-)certainty about predictions and should warn an explainee if the data point considered is `\emph{far away}' from the \emph{support} of the training set. 
    \item \textbf{Representativeness}: explanations should `\emph{cover}' training data evenly. This property is particularly relevant for prototype-based explanation techniques, \emph{e.g.} \cite{Kim2014} or \cite{Kim2016}, and, arguably, is of interest from the perspective of fairness (see Section~\ref{sec:intro}).
\end{itemize}

\subsubsection{Attribution Methods\label{sec:attrib}}

Attribution methods are, arguably, the commonest family of explanation techniques. Sundararajan \emph{et al.} \cite{Sundararajan2017} define an attribution as follows. For a function $f:\mathbb{R}^p\rightarrow[0,1]$ and an input $\matr{x}\in\mathbb{R}^p$, an \emph{attribution} for $\matr{x}$ w.r.t. some \emph{baseline} $\matr{x}_0$ is given by $A_f\left(\matr{x}, \matr{x}_0\right)=\begin{pmatrix}a_1 & \cdots & a_p\end{pmatrix}^\top\in\mathbb{R}^p$ wherein $a_i$ is the contribution of $x_i$ to the prediction made for $\matr{x}$ by $f(\cdot)$.\magic{attribution} Note that some of the methods described below do not require a reference sample $\matr{x}_0$. We will use this notation throughout the remainder of this section.

Ancona \emph{et al.} \cite{Ancona2019} distinguish two fundamentally different categories of attribution methods: \emph{sensitivity} methods which quantify how strongly the output of the function $f(\cdot)$ changes if an input variable is perturbed;\magic{sensitivity} and \emph{salience} methods which quantify marginal effects of variables on the output of $f(\cdot)$ compared to a baseline\footnote{For example, given by the output for the same input, but with the variable of interest masked or removed.}.\magic{salience} Most of the recent attribution techniques focus specifically on explaining (deep) neural network models. Many of these techniques implicitly or explicitly rely on \emph{gradient} information to produce attributions \cite{Ancona2019}. In the remainder of this section, we describe a few popular and archetypal examples of attribution techniques. 

\medskip

\textbf{LIME}.\quad Ribeiro \emph{et al.} \cite{Ribeiro2016} propose \emph{Local Interpretable Model-agnostic Explanations} (LIME) which identify interpretable data representations faithful to a given black box classifier. The explanation is defined by the following optimisation problem:
\begin{equation}
    \xi(\matr{x})=\arg\min_{g\in\mathfrak{G}}\mathcal{L}\left(f, g, \pi_{\matr{x}}\right)+\Omega(g),
    \label{eqn:lime}
\end{equation}
where $\mathfrak{G}$ is a class of explanation models; $\mathcal{L}(\cdot,\cdot,\cdot)$ is a \emph{fidelity function} which quantifies unfaithfulness of $g(\matr{x})$ in approximating $f(\cdot)$ within the \emph{proximity} of $\matr{x}$, defined by $\pi_{\matr{x}}$; and $\Omega(\cdot)$ is a complexity penalty. Essentially, $\mathcal{L}(\cdot,\cdot,\cdot)$ is a \emph{locality-aware} loss function and, in practice, is minimised in a model-agnostic manner. Usually $\mathfrak{G}$ is chosen to be a \emph{restricted} class of models that are intrinsically interpretable (see Section~\ref{sec:interpretablemodels}), \emph{e.g.} linear models, GAMs etc. During training, instances are sampled around each data point $\matr{x}_i$ weighted by $\pi_{\matr{x}_i}$. In addition to local explanations produced by $\xi(\cdot)$, Ribeiro \emph{et al.} \cite{Ribeiro2016} introduce a procedure for obtaining a global understanding of the model $f(\cdot)$. Given a limited budget, this procedure picks a number of explanations based on \emph{greedy submodular optimisation} \cite{Krause2014} and aggregates them into global variable importance statistics.

\medskip

\textbf{DeepLIFT}.\quad Shrikumar \emph{et al.} \cite{Shrikumar2017} take a different approach to attribution, introducing an efficient method for disentangling contributions of inputs in a neural network -- \emph{\underline{deep} \underline{l}earning \underline{i}mportant \underline{f}ea\underline{t}ures} (DeepLIFT). As opposed to LIME, DeepLIFT relies on comparisons to a \emph{reference} (baseline) data point. 
    
Let $t$ be the target neuron of interest (usually one of the output neurons) and let $\eta_1,\eta_2,...,\eta_K$ be intermediate neurons (potentially, from several layers) that suffice to compute $t$. Let $\Delta t=t-t_0$ be the difference from the reference. We then seek to assign contribution scores $C_{\Delta\eta_i\Delta t}$ so that they satisfy the \emph{summation-to-delta} property:
\begin{equation}
    \sum_{i=1}^KC_{\Delta\eta_i\Delta t}=\Delta t.
    \label{eqn:sumtodelta}
\end{equation}
An intuitive interpretation is that $C_{\Delta\eta_i\Delta t}$ is the amount of `blame' for the difference in outputs assigned to a difference in intermediate neurons. Importantly, $C_{\Delta\eta_i\Delta t}$ are not restricted to be $0$ when $\frac{\partial t}{\partial \eta_i}=0$ and, thus, arguably, address some of the limitations of gradient-based measures.
\newpage

Furthermore, by analogy to the partial derivative, Shrikumar \emph{et al.} \cite{Shrikumar2017} define the \emph{multiplier} as follows:
\begin{equation}
    m_{\Delta\eta_i\Delta t}=\frac{C_{\Delta\eta_i\Delta t}}{\Delta t}.
    \label{eqn:multiplier}
\end{equation}
In practice, we are not necessarily interested in contributions of hidden layer units. Therefore, we consider the following definition of multipliers for \emph{inputs}, which is consistent with the summation-to-delta property given by Equation~\ref{eqn:sumtodelta}:
\begin{equation}
    m_{\Delta x_i\Delta t}=\sum_{j}m_{\Delta x_i\Delta \eta_j}m_{\Delta\eta_j\Delta t},
    \label{eqn:chainrule}
\end{equation}
where $\eta_j$ denote hidden units. Equation~\ref{eqn:chainrule} is informally referred to as the \emph{chain rule for multipliers}. The authors propose several propagation rules for assigning $C_{\Delta\eta_i\Delta t}$ which alongside with the summation-to-delta and chain rule properties facilitate the attribution w.r.t. to an appropriate baseline sample. Key challenges in applying the DeepLIFT framework are adapting it to more complex or specialised neural network architectures and finding an informative reference sample (or multiple ones) \cite{Shrikumar2017}.

\medskip

\textbf{SHAP}.\quad Lundberg and Lee \cite{Lundberg2017} present a framework of \emph{\underline{Sh}apley \underline{a}dditive ex\underline{p}lanations} (SHAP). This framework builds on \emph{Shapley regression values}, inspired by the game theoretic concept of \emph{Shapley values} \cite{Hart1989}.\magic{Shapley values} For the $j$-th variable, the Shapley regression value is given by
\begin{equation}
    \phi_j=\sum_{\mathcal{S}\subseteq\mathcal{F}\setminus\left\{j\right\}}\frac{\left|\mathcal{S}\right|!\left(\left|\mathcal{F}\right|-\left|\mathcal{S}\right|-1\right)!}{\left|\mathcal{F}\right|!}\left\{f_{\mathcal{S}\cup\left\{j\right\}}\left(\matr{x}_{\mathcal{S}\cup\left\{j\right\}}\right)-f_{\mathcal{S}}\left(\matr{x}_{\mathcal{S}}\right)\right\},
    \label{eqn:shapregression}
\end{equation}
where $\mathcal{F}=\left\{1,2,...,p\right\}$ is the set of all predictor variables; $\matr{x}_{\mathcal{S}}$ is a vector composed of the components of $\matr{x}$ that are in $\mathcal{S}\subseteq\mathcal{F}$; and $f_{\mathcal{S}}(\cdot)$ is a function fitted in predictor variables in the set $\mathcal{S}$. In practice, Equation~\ref{eqn:shapregression} does not have to be evaluated exactly and can be approximated by randomly sampling subsets $\mathcal{S}\subseteq\mathcal{F}$.
    
Lundberg and Lee \cite{Lundberg2017} propose a model-agnostic kernel approximation of Shapley regression values described above. They also demonstrate that both LIME \cite{Ribeiro2016} and DeepLIFT \cite{Shrikumar2017} are special cases of the SHAP framework that resort to model-specific approximations of Equation~\ref{eqn:shapregression}.

\medskip    
    
\textbf{Integrated Gradients}.\quad Sundararajan \emph{et al.} \cite{Sundararajan2017} introduce the attribution method of \emph{integrated gradients} that is motivated by two fundamental axioms: \emph{sensitivity} axiom posits that (\emph{i}) if an input differs from a baseline in \emph{one} variable and has a prediction outcome different from the baseline, then the differing variable should be assigned with a non-zero attribution; (\emph{ii}) if $f(\cdot)$ is constant in some variable, then this variable should be given zero attribution; \emph{implementation invariance} states that for two \emph{functionally equivalent} black box models, attributions should be identical. Integrated gradients satisfy point (\emph{i}) of sensitivity and implementation invariance. For the $j$-th variable and baseline $\matr{x}_0$, the integrated gradient is given by 
\begin{equation}
    \text{IG}_j^f(\matr{x})=(x_j-x_{0j})\int_{\alpha=0}^1\frac{\partial f\left(\matr{x}_0+\alpha[\matr{x}-\matr{x}_0]\right)}{\partial x_j}d\alpha.
    \label{eqn:integrad}
\end{equation}
Observe that $\text{IG}^f(\matr{x})$ is an integral\footnote{In practice, evaluated using the Riemann approximation.} of gradients along the \emph{straight} path between $\matr{x}$ and $\matr{x}_0$.\magic{integrated gradient} Yet another important property satisfied by integrated gradients is \emph{completeness}: if $f$ is differentiable almost everywhere, $\sum_{j=1}^p\text{IG}_j^f(\matr{x})=f(\matr{x})-f(\matr{x}_0)$.
    
Definition in Equation~\ref{eqn:integrad} can be generalised further by considering some non-straight path between $\matr{x}$ and $\matr{x}_0$. \emph{Path integrated gradients} are defined for specified \emph{paths} $\gamma=\left(\gamma_1,...,\gamma_p\right):[0,1]\rightarrow\mathbb{R}^p$:\magic{path integrated gradient}
\begin{equation}
    \text{IG}_j^{f\gamma}(\matr{x})=\int_{\alpha=0}^1\frac{\partial f\left(\gamma(\alpha)\right)}{\partial\gamma_j(\alpha)}\frac{\partial\gamma_j(\alpha)}{\partial\alpha}d\alpha.
    \label{eqn:pathintegrad}
\end{equation}
Path integrated gradients are \emph{unique} attributions that fulfil points (\emph{i}) and (\emph{ii}) of sensitivity, implementation invariance, and completeness. In fact, path integrated gradients correspond to a generalisation of \emph{Shapley values} proposed by Aumann and Shapley \cite{Aumann1974} in the context of \emph{infinite games} within the \emph{cooperative game theory}.\magic{Aumann-Shapley values} 
    
Choosing paths of integration in an informed manner has been a topic of fruitful research, \emph{e.g.} Jha \emph{et al.} \cite{Jha2020} propose \emph{enhanced} integrated gradients and demonstrate that nonlinear paths can improve interpretability on the case study of RNA splicing prediction. A noteworthy extension of integrated gradients is discussed in \cite{Dhamdhere2018} by Dhamdhere \emph{et al.} who introduce a notion of \emph{conductance} which quantifies a `\emph{flow}' of attribution through an individual hidden unit. Merrill \emph{et al.} \cite{Merrill2019} further extend integrated gradients to a wider range of model classes by considering \emph{partially} differentiable functions.
    
\medskip

\textbf{Full-gradient Representations of Neural Networks}.\quad Srinivas and Fleuret \cite{Srinivas2019} attempt to reconcile local and global perspectives on variable importance and introduce \emph{full-gradient} (\mbox{FullGrad}) saliency maps which assign importance to both input variables and \emph{feature detectors} (intermediate units) within a neural network. The framework is motivated by two \emph{relaxed} notions of local and global attribution: \emph{weak dependence} and \emph{completeness}. 
    
A \emph{saliency map} is a function $S(\matr{x})=\sigma\left(f,\matr{x}\right)\in\mathbb{R}^p$.\magic{saliency maps} $S(\cdot)$ is said to be \emph{weakly dependent} if for a piecewise linear function $$f(\matr{x})=\begin{cases}
    \matr{w}_0^\top\matr{x}+b_0\text{, }\matr{x}\in\mathcal{U}_0\\
    \vdots\\
    \matr{w}_B^\top\matr{x}+b_B\text{, }\matr{x}\in\mathcal{U}_B,
\end{cases},$$
where $\mathcal{U}_b\subset\mathbb{R}^p$, $0\leq b\leq B$, are open and connected sets, $S(\matr{x})$ is constant in $\matr{x}$ within $\mathcal{U}_b$, for all $b$. According to \cite{Srinivas2019}, $S(\cdot)$ is \emph{complete}\footnote{In addition, Srinivas and Fleuret \cite{Srinivas2019} define a notion of \emph{completeness w.r.t. to a baseline}, similarly to \cite{Sundararajan2017}.} if there exists a function $\phi(\cdot,\cdot)$ s.t. $\phi(S(\matr{x}),\matr{x})=f(\matr{x})$, for all $f(\cdot)$ and $\matr{x}$. It can be shown that, for a general piecewise linear $f(\cdot)$, there exists no saliency map $S(\cdot)$ that is both weakly dependent and complete.
    
For a neural network with \emph{ReLU} activations and biases, \emph{implicit and explicit}, $\matr{b}\in\mathbb{R}^F$, the full-gradient representation is specified by 
\begin{equation}
    f\left(\matr{x};\matr{b}\right)=\nabla_{\matr{x}}f(\matr{x};\matr{b})^\top\matr{x} + \nabla_{\matr{b}}f(\matr{x};\matr{b})^\top\matr{b}.
    \label{eqn:fullgrad}
\end{equation}
Note that for non-ReLU networks this representation can be approximated by linearising activation functions. A full gradient is, generally, given by a pair $G=\left(\nabla_{\matr{x}}f(\matr{x}),f^{\matr{b}}(\matr{x})\right)\in\mathbb{R}^{p+F}$.\magic{full gradient} If the neural network contains only ReLU nonlinearities without batch-norm, then $G$ is weakly dependent and complete, thus, striking a perfect balance between locality and globality. In addition to he theoretical result above, Srinivas and Fleuret \cite{Srinivas2019} provide a practical approximation for the full-gradient representation of convolutional neural networks.

\bigskip

Although attribution methods have become an active and important area of research, their general applicability and usefulness have been scrutinised and criticised. For instance, Rudin \cite{Rudin2019} argues that explanation and, in particular, attribution methods cannot be completely faithful w.r.t. the original black box model and that attributions often do not provide sufficient information about \emph{how} the model \emph{actually} works, they rather tell us where the model \emph{looks}. Kumar \emph{et al.} \cite{Kumar2020} criticise Shapley-value-based explanations, such as described in \cite{Lundberg2017, Shrikumar2017, Ribeiro2016}, for their reliance on the \emph{additivity axiom} \cite{Hart1989} and lack of human-groundedness and contrastiveness.

\subsubsection{Symbolic Metamodels\label{sec:symmeta}}

In Section~\ref{sec:symreg}, we described symbolic regression as an approach to learning interpretable mathematical expressions from raw data. Similar to LIME \cite{Ribeiro2016} (see Section~\ref{sec:attrib}), symbolic regression can be used for \emph{metamodelling} already learnt opaque predictive models. To this end, Alaa and van der Schaar \cite{Alaa2019} propose an elegant parameterisation of the symbolic regression problem that admits optimisation by gradient descent, in contrast to genetic programming \cite{AmirHaeri2017} and simulated annealing \cite{Stinstra2007} techniques which require searching through a \textit{discrete} solution space.

According to Alaa and van der Schaar \cite{Alaa2019}, the \emph{symbolic metamodelling problem} reduces to the following (\emph{cf.} Equation~\ref{eqn:symreg}):\magic{symbolic metamodelling}
\begin{equation}
    \min_{g\in\mathfrak{G}}\ell\left(g, f\right)=\min_{g\in\mathfrak{G}}\int_{\mathcal{X}}\left(g(\matr{x})-f(\matr{x})\right)^2d\matr{x},
    \label{eqn:metamodel}
\end{equation}
where $\ell(\cdot,\cdot)$ is a metamodelling loss and $\mathfrak{G}$ is a class of \emph{succinct} mathematical expressions. In this section, we assume $D$-dimensional data points, \emph{i.e.} $\matr{x}\in\mathbb{R}^D$. We seek a parameterisation of $\mathfrak{G}$ which would make the optimisation problem in Equation~\ref{eqn:metamodel} `\emph{easier}'; in particular, given parameterisation $\mathfrak{G}=\left\{G(\cdot,\theta):\theta\in\Theta\right\}$, the problem above reduces to $\min_{\theta\in\Theta}\ell\left(G(\cdot,\theta), f(\cdot)\right)$. 

By Kolmogorov--Arnold \emph{superposition theorem} \cite{Kolmogorov1956, Arnold1957, Alaa2019}, function $g(\cdot)$ can be rewritten in the following form:
\begin{equation}
    g(\matr{x})=\sum_{i=0}^rg_i^{\text{out}}\left(\sum_{j=1}^Dg_{i,j}^{\text{in}}(x_j)\right),
    \label{eqn:superposition}
\end{equation}
where $g_{i,j}^{\text{in}}(\cdot)$ and $g_i^{\text{out}}(\cdot)$ are continuous \emph{basis functions}. Observe that if $r=1$, then $g(\cdot)$ is a GAM \cite{Hastie1986} (see Section~\ref{sec:gams}). Equation~\ref{eqn:superposition} yields parameterisation $G(\matr{x};\theta)=G\left(\matr{x};\left\{g_{i,j}^{\text{in}}\right\}_{i,j},\left\{g_i^{\text{out}}\right\}_i\right)$. Alaa and van der Schaar \cite{Alaa2019} further parameterise basis functions themselves by representing them as \emph{Meijer G-functions} \cite{Meijer1936},\magic{Meijer G-functions} \emph{i.e.} in the form of a linear integral in complex plane $s$:
\begin{equation}
    G^{m,n}_{p,q}\left(\begin{matrix}a_1,a_2,...,a_p \\ b_1,b_2,...,b_q\end{matrix}\middle|x\right)=\frac{1}{2\pi i}\int_{\mathcal{L}}\frac{\prod_{j=1}^{m}\Gamma(b_j-s)\prod_{j=1}^n\Gamma(1-a_j+s)}{\prod_{j=m+1}^q\Gamma(1-b_j+s)\prod_{j=n+1}^p\Gamma(a_j+s)}x^sds,
    \label{eqn:meijergfunc}
\end{equation}
where $\Gamma(\cdot)$ is the Gamma function; $\mathcal{L}$ is the integration path; $\matr{a}_p=\begin{pmatrix}a_1 & a_2 & ... & a_p\end{pmatrix}$ and $\matr{b}_q=\begin{pmatrix}b_1 & b_2 & ... & b_q\end{pmatrix}$ are real-valued parameters that, alongside with $m$ and $n$, determine poles and zeros of the integrand. Most familiar functional forms are special cases of Meijer G-functions. In the end, parameterisation $G(\matr{x};\theta)$ is given by:
\begin{equation}
    G(\matr{x};\theta)=\sum_{i=0}^rG^{m,n}_{p,q}\left(\theta^{\text{out}}_i\middle|\sum_{j=1}^DG_{p,q}^{m,n}\left(\theta_{i,j}^{\text{in}}\middle|x_j\right)\right).
    \label{eqn:parameterisation}
\end{equation}
An important property of Meijer G-functions is their closure under differentiation. This allows searching $\mathfrak{G}$ efficiently using the gradient descent procedure.

Symbolic metamodelling \cite{Alaa2019} and other symbolic regression techniques \cite{Stinstra2007, Jin2019, Udrescu2020} are a compelling alternative to attribution methods (see Section~\ref{sec:attrib}), especially, when a parsimonious analytical representation of the learnt function is desirable. The parameterisation proposed by Alaaa and van der Schaar \cite{Alaa2019} is an interesting and promising reformulation of this classical problem.

\subsubsection{Counterfactual Explanations\label{sec:counterfactuals}}

In some applications, it might be of key interest to provide \emph{human-friendly} explanations \cite{Carvalho2019} which are comprehensible to a wide nonspecialist audience. Techniques discussed so far were mostly addressing the question ``\emph{Why this prediction was made?}'', \emph{counterfactual explanations}, on the other hand, try to answer the question ``\emph{Why this prediction was made instead of another one?}'' \cite{Carvalho2019}.\magic{counterfactual explanations} Such techniques focus on producing \emph{contrastive} and \emph{actionable local} explanations, which can be useful in various real-world settings, \emph{e.g.} when suggesting lifestyle changes to a patient to reduce her risks or providing reasons for the low creditworthiness of a company.

Wachter \emph{et al.} \cite{Wachter2017} formalise counterfactual explanations in the context of explainable ML. They propose solving the following optimisation problem to find a counterfactual explanation $\matr{x}'$ for a data point $\left(\matr{x}_i,y_i\right)$ and a black box model $f(\cdot)$:
\begin{equation}
    \min_{\matr{x}'}d(\matr{x}_i,\matr{x}')+\lambda\ell\left(f(\matr{x}'),y'\right),
    \label{eqn:cfwachter}
\end{equation}
where $d(\cdot,\cdot)$ is an appropriate distance function; $y'$ is chosen to be meaningfully different from $y_i$ and $\ell(f(\matr{x}'),y')$ quantifies how `different' output for $\matr{x}'$ is from the $y'$ chosen, \emph{e.g.} MSE for regression or hinge loss for classification; $\lambda$ is a parameter controlling the slackness on the constraint $f(\matr{x}')=y'$. The problem given by Equation~\ref{eqn:cfwachter} is reminiscent of generating \emph{adversarial perturbations}. Perturbations are encouraged to be sparse by penalising $d(\matr{x}_i,\matr{x}')$. In particular, Wachter \emph{et al.} propose using the following distance function:
\begin{equation}
    d(\matr{x},\matr{x}')=\sum_{j=1}^p\frac{\left|x_j-x'_j\right|}{\text{MAD}_j},
    \label{eqn:distwachter}
\end{equation}
where $\text{MAD}_j$ denotes the median absolute deviation of the $j$-th variable. Distance function in Equation~\ref{eqn:distwachter} favours sparse changes in $\matr{x}'$ compared to $\matr{x}$ and accounts for variability of individual features. 

Mothilal \emph{et al.} \cite{Mothilal2020} extend the framework introduced by Wachter \emph{et al.} \cite{Wachter2017} focusing on providing \emph{multiple diverse} counterfactual explanations.\magic{diversity} In particular, for a data point $\matr{x}_i$, explanations $\matr{c}_1,...,\matr{c}_K$ are found by solving the optimisation problem below (\emph{cf.} Equation~\ref{eqn:cfwachter}):
\begin{equation}
    \min_{\matr{c}_1,...,\matr{c}_K}\frac{1}{K}\sum_{k=1}^K\ell\left(f(\matr{c}_k),y'\right)+\frac{\lambda_1}{K}\sum_{k=1}^Kd\left(\matr{x}_i,\matr{c}_k\right)-\lambda_2\det\left(\matr{S}\right),
    \label{eqn:cfmothilal}
\end{equation}
where $S_{k,l}=\frac{1}{1+d\left(\matr{c}_k,\matr{c}_l\right)}$, and $\det\left(\matr{S}\right)$ quantifies diversity among  counterfactual explanations $\matr{c}_1,...,\matr{c}_K$. In addition, Mothilal \emph{et al.} \cite{Mothilal2020} propose \emph{quantitative evaluation metrics} for counterfactual explanation techniques.\magic{evaluation of counterfactual explanations} \emph{Validity} quantifies how many of the proposed explanations are actually counterfactual. In the setting of classification, validity is given by
\begin{equation}
    \frac{1}{K}\sum_{k=1}^K\mathbbm{1}_{\left\{f(\matr{c}_k)\neq y_i\right\}}.
    \label{eqn:cfvalidity}
\end{equation}
\emph{Proximity} measures the `closeness' of explanations to the original data point:
\begin{equation}
    -\frac{1}{K}\sum_{k=1}^Kd\left(\matr{c}_k,\matr{x}_i\right).
    \label{eqn:cfproximity}
\end{equation}
\emph{Sparsity} quantifies how sparse are the perturbations of $\matr{x}_i$ induced by the set of explanations:
\begin{equation}
    1-\frac{1}{Kp}\sum_{k=1}^K\sum_{j=1}^p\mathbbm{1}_{\left\{c_{k,j}\neq x_{i,j}\right\}}.
    \label{eqn:cfsparsity}
\end{equation}
Finally, \emph{diversity} evaluates how diverse proposed explanations are:
\begin{equation}
    \Delta\left(\matr{c}_1,...,\matr{c}_K\right)=\frac{1}{K^2}\sum_{k=1}^{K-1}\sum_{l=k+1}^{K}d\left(\matr{c}_k,\matr{c}_l\right).
    \label{eqn:cfdiversity}
\end{equation}

Approaches of Wachter \emph{et al.} \cite{Wachter2017} and Mothilal \emph{et al.} \cite{Mothilal2020} rely on performing gradient descent and, thus, assume that the black box model $f(\cdot)$ is differentiable. Karimi \emph{et al.} \cite{Karimi2020} generalise this framework and propose \emph{model-agnostic counterfactual explanations} (MACE). They transform the original optimisation problem into a sequence of Boolean satisfiability problems (SAT) and leverage powerful satisfiability modulo theories (SMT) solvers to solve these. An important advantage of MACE is its complete agnosticism to the choice of the black box model or distance function and ability to incorporate \emph{plausibility constraints} that allow injecting domain-specific knowledge.\magic{plausibility constraints}

The problem of counterfactual explanation (see Equations~\ref{eqn:cfwachter} and \ref{eqn:cfmothilal}) naturally admits a \emph{generative model} as a solution. Recently, several papers have utilised deep generative models \cite{Chang2019, Mahajan2019, Liu2019} to address problems similar to the ones considered by Wachter \textit{et al.} \cite{Wachter2017} and Mothilal \emph{et al.} \cite{Mothilal2020}. Chang \emph{et al.} \cite{Chang2019} introduce saliency maps based on \emph{counterfactual generation} with masking for explaining image classifiers.\magic{counterfactual generation} Liu \emph{et al.} \cite{Liu2019} leverage GANs to generate minimal change counterfactual examples for image classifiers. Last but not least, Mahajan \emph{et al.} propose a VAE-based counterfactual generative model that focuses on feasibility and preservation of \emph{causal} constraints, introducing regularisation derived from a \emph{structural causal model} (SCM) \cite{Pearl2010}.

Another perspective on counterfactual explanations is provided by \emph{algorithmic recourse}, surveyed in detail by Karimi \emph{et al.} in \cite{Karimi2020c}. Algorithmic recourse focuses on explaining the decisions and recommending further actions to ``\emph{individuals who are unfavourably treated by automated decision-making systems}'' \cite{Karimi2020c}. In \cite{Karimi2020b}, Karimi \emph{et al.} criticise counterfactual explanations for the lack of actionability and provide a causal perspective of algorithmic recourse by considering \emph{interventions} rather than explanations. To avoid infeasible or costly recommendations resulting from na\"{i}ve counterfactual explanations \cite{Wachter2017, Mothilal2020} Karimi \emph{et al.} \cite{Karimi2020b} propose finding minimal cost structural interventions resulting in a favourable outcome. While this approach certainly offers an interesting and, possibly, more user-centred perspective, the core limitation of algorithmic recourse, as outlined in \cite{Karimi2020b} and \cite{Karimi2020d}, is the unrealistic assumption of a known causal structure.
\newpage

\begin{table}[H]
    \caption{Properties of reviewed interpretable models. ``$\CIRCLE$'' and ``$\odot$'' denote globally- and locally-interpretable methods, respectively. ``\Checkmark'' signifies that a property (columns) is satisfied by a technique (rows). ``$\boldsymbol{\sim}$'' denotes that a property either holds partially or that a method could be easily extended/adjusted to satisfy the property (see footnotes for detailed explanations).}
    \label{tab:interpretmethodsummary}
    \footnotesize
        \begin{center}
            \begin{tabular}{l||ccccccc}
                \hline
                \textbf{Method} & \textbf{Scale} & \textbf{Linearity} & \textbf{Sparsity} & \textbf{Additivity} & \textbf{Monotonicity} & \makecell{\textbf{Unstructured} \\ \textbf{Data}} \\
                \hline
                \makecell[l]{FLRs \\ \cite{Wang2015, Wang2015b, Chen2018}} & $\CIRCLE$ &  & \Checkmark &  & \Checkmark &  \\
                \hline
                \makecell[l]{SLIMs \\ \cite{Ustun2015, Ustun2017}} & $\CIRCLE$ & \Checkmark & \Checkmark & \Checkmark &  & \\
                \hline
                \makecell[l]{GAMs \\ \cite{Hastie1986, Lou2012}} & $\CIRCLE$ &  &  & \Checkmark &  &  \\
                \hline
                \makecell[l]{GA\textsuperscript{2}Ms \\ \cite{Caruana2015}} & $\CIRCLE$ &  &  & $\boldsymbol{\sim}$\textsuperscript{a} &  &  \\
                \hline
                \makecell[l]{SpAMs \\ \cite{Ravikumar2007}} & $\CIRCLE$ &  & \Checkmark & \Checkmark &  &  \\
                \hline
                \makecell[l]{SPINNs \\ \cite{Feng2017, Tank2018, Khanna2020}} & $\CIRCLE$ &  & \Checkmark &  &  &  \Checkmark \\
                \hline
                \makecell[l]{DeepPINK \\ \cite{Lu2018}} & $\CIRCLE$ &  & \Checkmark &  &  &  \Checkmark \\
                \hline
                \makecell[l]{VCMs \\ \cite{Hastie1993}} & $\odot$ &  &  &  &  \\
                \hline
                \makecell[l]{CENs \\ \cite{AlShedivat2017}} & $\odot$ &  &  &  &  & \Checkmark \\
                \hline
                \makecell[l]{SENNs \\ \cite{AlvarezMelis2018, Quinn2020}} & $\odot$ &  &  &  &  &  \Checkmark \\
                \hline
                \makecell[l]{AMEs \\ \cite{Schwab2019}} & $\odot$ &  &  &  &  &  \Checkmark \\
                \hline
                \makecell[l]{Symbolic \\ regression \\ \cite{AmirHaeri2017, Jin2019, Udrescu2020}} & $\CIRCLE$ & $\boldsymbol{\sim}$\textsuperscript{b} & $\boldsymbol{\sim}$\textsuperscript{b} & $\boldsymbol{\sim}$\textsuperscript{b} & $\boldsymbol{\sim}$\textsuperscript{b} &  \\
                \hline
            \end{tabular}
        \end{center}
        {\scriptsize
        \textsuperscript{a} GA\textsuperscript{2}Ms \cite{Caruana2015} are not \emph{completely} additively separable: they contain pairwise interaction terms.
        
        \textsuperscript{b} Properties of symbolic regression models \cite{AmirHaeri2017, Jin2019, Udrescu2020} depend on the choice of the hypothesis class $\mathfrak{G}$ (see Equation~\ref{eqn:symreg}).}
\end{table}

\begin{table}[H]
    \caption{Properties of reviewed explanation techniques. ``$\CIRCLE$'' and ``$\odot$'' denote global and local explanation methods, respectively. ``\Checkmark'' signifies that a property (columns) is satisfied by a technique (rows). ``$\boldsymbol{\sim}$'' denotes that a property either holds partially or that a method could be easily extended/adjusted to satisfy the property (see footnotes for detailed explanations).}
    \label{tab:explmethodsummary}
    \footnotesize
        \begin{center}
            \begin{tabular}{l||cccccccc}
                \hline
                \textbf{Method} & \textbf{Scale} & \makecell[c]{\textbf{Attri-}\\\textbf{bution}} & \textbf{Agnosticism} & \makecell[c]{\textbf{Reference-} \\ \textbf{free}} & \makecell[c]{\textbf{Contras-} \\ \textbf{tiveness}} & \textbf{Diversity} & \textbf{Causality} & \makecell{\textbf{Unstructured} \\ \textbf{Data}} \\
                \hline
                \makecell[l]{LIME \\ \cite{Ribeiro2016}} & $\odot$,$\CIRCLE$ & \Checkmark & \Checkmark & \Checkmark &  &  &  & \Checkmark \\
                \hline
                \makecell[l]{DeepLIFT \\ \cite{Shrikumar2017}} & $\odot$ & \Checkmark &   &  &  &  &  & \Checkmark \\
                \hline
                \makecell[l]{SHAP \\ \cite{Lundberg2017}} & $\odot$ & \Checkmark & \Checkmark & \Checkmark &  &  &  & \Checkmark \\
                \hline
                \makecell[l]{IG \\ \cite{Sundararajan2017, Jha2020, Merrill2019}} & $\odot$ & \Checkmark &  &  &  &  &  & \Checkmark \\
                \hline
                \makecell[l]{Full Gradient \\ \cite{Srinivas2019}} & $\odot,\CIRCLE$ & \Checkmark &  & \Checkmark &  &  &  & \Checkmark \\
                \hline
                \makecell[l]{Symbolic \\ Metamodels \\ \cite{Alaa2019}} & $\CIRCLE$ & \Checkmark & \Checkmark & \Checkmark &  &  &  &  \\
                \hline
                \makecell[l]{CF Explanations \\ \cite{Wachter2017}} & $\odot$ &  &  & \Checkmark & \Checkmark &  &  & $\boldsymbol{\sim}$\textsuperscript{a} \\
                \hline
                \makecell[l]{Diverse CF \\ Explanations \\ \cite{Mothilal2020}} & $\odot$ &  &  & \Checkmark & \Checkmark & \Checkmark &  & $\boldsymbol{\sim}$\textsuperscript{a} \\
                \hline
                \makecell[l]{MACE \\ \cite{Karimi2020}} & $\odot$ &  & \Checkmark & \Checkmark & \Checkmark &  &  & $\boldsymbol{\sim}$\textsuperscript{a} \\
                \hline
                \makecell[l]{VAE-based CF \\ Explanations \\ \cite{Mahajan2019}} & $\odot$ &  &  \Checkmark & \Checkmark & \Checkmark & $\boldsymbol{\sim}$\textsuperscript{b} & $\boldsymbol{\sim}$\textsuperscript{c} & \Checkmark\\
                \hline
                \makecell[l]{GAN-based CF \\ Explanations \\ \cite{Liu2019}} & $\odot$ &  &  \Checkmark & \Checkmark & \Checkmark & $\boldsymbol{\sim}$\textsuperscript{b} &  & \Checkmark\\
                \hline
                \makecell[l]{FIDO CF \\ Saliency Maps \\ \cite{Chang2019}} & $\odot$ & \Checkmark & \Checkmark & \Checkmark & $\boldsymbol{\sim}$\textsuperscript{d} &  &  & \Checkmark\\
                \hline
                \makecell[l]{Interventions for \\ Algorithmic Recourse \\ \cite{Karimi2020c, Karimi2020d}} & $\odot$ &  &  & \Checkmark & \Checkmark &  & \Checkmark & \\
                \hline
            \end{tabular}
        \end{center}
        {\scriptsize
        \textsuperscript{a} Although methods proposed in \cite{Wachter2017, Mothilal2020, Karimi2020} are only tested on tabular datasets, their formulation, in principle, generalises to some types of unstructured data, \emph{e.g.} images.
        
        \textsuperscript{b} Although VAE- and GAN-based counterfactual explanations \cite{Mahajan2019, Liu2019} do not explicitly maximise diversity, the underlying generative models tend to produce diverse samples.
        
        \textsuperscript{c} Causal constraints can be incorporated into the framework proposed by Mahajan \emph{et al.} \cite{Mahajan2019} via regularisation or user feedback.
        
        \textsuperscript{d} In \cite{Chang2019}, saliency maps are aggregated across generated counterfactual samples, thus, contrastiveness is not explicit as in other CF explanation techniques.}
\end{table}
\newpage

\section{Summary\label{sec:summary}}

In this literature review, we provided a survey of interpretable and explainable machine learning methods (see Tables~\ref{tab:interpretmethodsummary} and \ref{tab:explmethodsummary} for the summary of the techniques), described commonest goals and desiderata for these techniques, motivated their relevance in several fields of application, and discussed their quantitative evaluation. Interpretability and explainability still remain an active area of research, especially, in the face of recent rapid progress in designing highly performant predictive models and inevitable infusion of machine learning into other domains, where decisions have far-reaching consequences.

For years the field has been challenged by a lack of clear definitions for interpretability or explainability, these terms being often wielded ``\emph{in a quasi-mathematical way}'' \cite{Lipton2018, Lipton2017}. For many techniques, there still exist no satisfactory functionally-grounded evaluation criteria and universally accepted benchmarks, hindering reproducibility and model comparison. Moreover, meaningful adaptations of these methods to `real-world' machine learning systems and data analysis problems largely remain a matter for the future. It has been argued that, for successful and widespread use of interpretable and explainable machine learning models, stakeholders need to be involved in the discussion \cite{Lipton2017, Rudin2019}. A meaningful and equal collaboration between machine learning researchers and stakeholders from various domains, such as medicine, natural sciences, and law, is a logical next step within the evolution of interpretable and explainable ML. 

\newpage

\bibliographystyle{ieeetr}
{\footnotesize
\bibliography{bibliography}
}

\newpage

\printindex

\end{document}